\pdfoutput=1

\documentclass[11pt]{article}

\usepackage[preprint]{acl}

\usepackage{times}
\usepackage{latexsym}
\usepackage{amsmath}
\usepackage{amssymb}
\usepackage{mathtools}
\usepackage{amsthm}
\usepackage{graphicx}
\usepackage{subfigure}
\usepackage{algorithm}
\usepackage{algorithmic}
\usepackage{booktabs}
\usepackage{multirow}
\usepackage{color,xcolor}
\usepackage{tcolorbox}
\usepackage{array}
\usepackage{ragged2e}
\usepackage[normalem]{ulem}
\useunder{\uline}{\ul}{}
\usepackage{wrapfig}

\usepackage[T1]{fontenc}

\usepackage[utf8]{inputenc}

\usepackage{microtype}

\usepackage{inconsolata}

\usepackage{graphicx}

\title{Refining Sentence Embedding Model through Ranking Sentences Generation with Large Language Models}

\author{ \textbf{Liyang He\textsuperscript{1}, Chenglong Liu\textsuperscript{1}, Rui Li\textsuperscript{1}, Zhenya Huang\textsuperscript{1,2}, Shulan Ruan\textsuperscript{3} }\\
\textbf{Jun Zhou\textsuperscript{4}, Enhong Chen\textsuperscript{1}}\thanks{Corresponding author} \\
\textsuperscript{1}State Key Laboratory of Cognitive Intelligence, University of Science and Technology of China; \\
\textsuperscript{2} Institute of Artificial Intelligence, Hefei Comprehensive National Science Center;  \\
\textsuperscript{3}Shenzhen International Graduate School, Tsinghua University; \\
\textsuperscript{4}Zhejiang University \\
\texttt{\{heliyang,sepcnt,ruili2000\}@mail.ustc.edu.cn},\texttt{slruan@sz.tsinghua.edu.cn} \\ \texttt{\{huangzhy,  cheneh\}@ustc.edu.cn}, \texttt{junzhougucas@gmail.com}}

\begin{document}
\maketitle
\begin{abstract}
Sentence embedding is essential for many NLP tasks, with contrastive learning methods achieving strong performance using annotated datasets like NLI. Yet, the reliance on manual labels limits scalability. Recent studies leverage large language models (LLMs) to generate sentence pairs, reducing annotation dependency. However, they overlook ranking information crucial for fine-grained semantic distinctions. To tackle this challenge, we propose a method for controlling the generation direction of LLMs in the latent space. Unlike unconstrained generation, the controlled approach ensures meaningful semantic divergence. Then, we refine exist sentence embedding model by integrating ranking information and semantic information. Experiments on multiple benchmarks demonstrate that our method achieves new SOTA performance with a modest cost in ranking sentence synthesis\footnote{Our code is available at \url{https://github.com/hly1998/RankingSentenceGeneration}}.
\end{abstract}

\section{Introduction}

Sentence embedding is a fundamental task in natural language processing. It provides effective semantic representations for various downstream applications, such as semantic search \cite{he2023efficient,li2024consider}, text classification \cite{wang2022sentence}, question-answering systems \cite{nguyen2022spbertqa}, and other specialized domain systems \cite{liu2019ekt,xue2024decompose}. In recent years, significant progress has been made in the study of sentence embeddings, with methods based on contrastive learning standing out in particular. These approaches learn embeddings of sentences by bringing semantically similar sentences closer together and pushing dissimilar ones further apart. Current mainstream research relies on high-quality annotated data, especially natural language inference (NLI) datasets \cite{bowman2015large,williams2018broad}. For instance, supervised contrastive learning methods based on NLI have demonstrated a remarkable ability to surpass the unsupervised approaches \cite{limkonchotiwat2022congen,jiang2022promptbert}. However, such annotated datasets are often unavailable in most real-world scenarios, and the manual construction of these datasets incurs extremely high costs.

\begin{figure*}[t]
	\centering
	\includegraphics[width=0.98\textwidth]{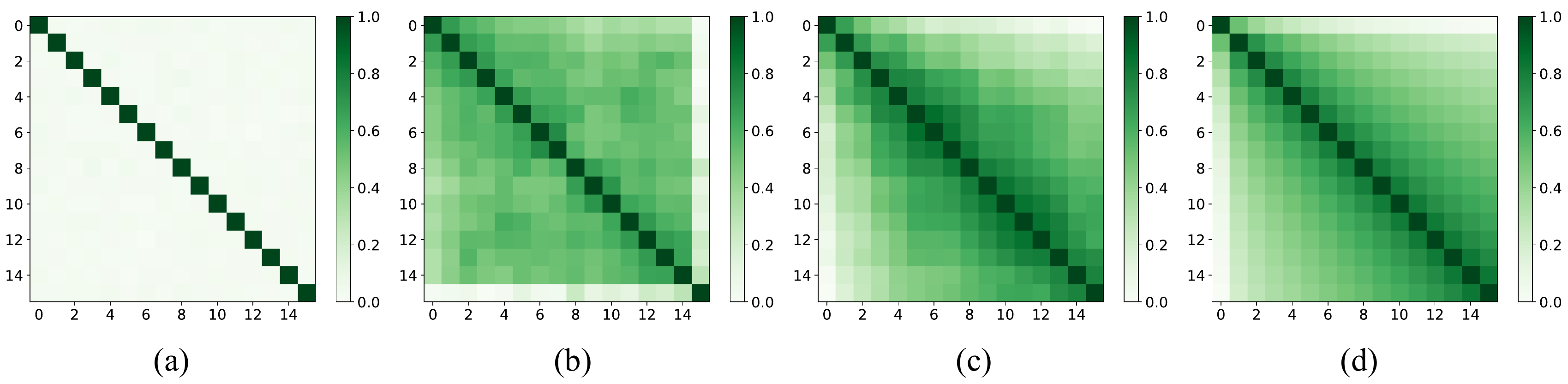}
	\caption{Sentence similarity within the ranking sentences obtained through different methods. We randomly selected 1,000 ranking sentences generated by these methods and extracted the first 16 sentences from each ranking sentence. Then, we use a trained DiffCSE \cite{chuang2022diffcse} to obtain their embeddings and compute their average similarity. (a) Directly extracted from the trained batch. (b) Prompting the LLM to generate complete ranking sentences at once. (c) Prompting the LLM to generate ranking sentences step by step. (d) Generating the ranking sentences using our proposed directionally controlled generation method.}
\label{fig:heat_map}
\end{figure*}


To reduce reliance on manually annotated data, recent studies have begun to explore leveraging the powerful generative capabilities of large language models (LLMs) to construct high-quality sentence pairs automatically. For instance, SynCSE \cite{zhang2023contrastive} employs LLMs to generate semantically similar sentence pairs, enhancing the effectiveness of contrastive learning. MultiCSR \cite{wang2024large} further evaluates the quality of LLM-generated outputs, filtering out erroneous results. GCSE \cite{lai2024enhancing} utilizes knowledge graphs to extract entities and quantities, enabling LLMs to generate more diverse and knowledge-enriched samples. These approaches significantly diminish the dependence on manual annotation.

However, current research focuses on generating sentence pairs, overlooking the critical role of ranking sentences. While sentence pairs can capture the similarity between sentences, they fail to effectively distinguish between ``highly similar'' sentences and ``slightly different''. \citet{liu2023rankcse} point out that the limitation of sentence pairs lies in their inability to represent finer-grained semantic distinctions. Existing unsupervised methods, such as RankCSE \cite{liu2023rankcse} and RankEncoder \cite{seonwoo2023ranking}, attempt to construct ranking information using in-batch data to address this shortcoming. However, the ranking information in these methods is derived in an unsupervised manner, lacking explicit ranking supervision signals. As shown in Figure~\ref{fig:heat_map} (a), the heatmap illustrates the similarity calculations between sentences acquired from the in-batch data. We observe that the relationships among sentences within the in-batch data are treated as equivalent, failing to capture the hierarchical semantic distinctions. Thus, we propose a new research question: \textbf{Can LLMs be used to generate ranking sentences to enhance the performance of sentence embedding models?}



A straightforward method for generating ranking sentences is to directly prompt LLMs to produce them. However, such an unconstrained generation process will result in ambiguous sentence semantic relationships. As illustrated in Figure~\ref{fig:heat_map} (b) and (c), neither prompting the LLM to generate complete ranking sentences at once nor guiding it to generate them step by step can ensure a gradual increase in semantic distance \footnote{We provide a detailed description of their generation process in Appendix~\ref{apd:generation}.}. Thus, it fails to provide high-quality ranking information for sentence embedding models.

In this paper, we propose a latent space directional control method for ranking sentence generation and a post-training method for synthesized ranking sentences. Specifically, we design a directionally controlled generation method that LLMs to produce ranking sentences. By utilizing the generation probabilities of the preceding two sentences, we ensure that the resulting latent space remains in a consistent direction. As shown in Figure~\ref{fig:heat_map} (d), our generated ranking sentences exhibit a gradual increase in semantic divergence within the semantic space. Then, we integrate the ranking information and semantic information from the synthesized ranking sentences to refine existing sentence embedding models through post-training. The contributions of this paper can be summarized as follows:
\begin{itemize}
\item We are the first to use LLMs to generate ranking sentences. We have curated a dataset consisting of 16,063 ranking sentences and 530,079 sentences, opening new avenues for research in sentence embedding.
\item We propose a post-training approach that incorporates both ranking and semantic information from the synthesized ranking sentences, substantially enhancing the performance of sentence embedding models on STS, reranking, and TR tasks.
\item Extensive experiments on multiple benchmark datasets demonstrate the effectiveness of the proposed method. Even using merely 5\% of the synthesized ranking sentences is sufficient to surpass the original sentence embedding model significantly.
\end{itemize}

\section{Background}

In unsupervised sentence embedding models, a series of works represented by SimCSE \cite{gao2021simcse} employ contrastive learning to acquire effective embeddings by bringing semantically similar neighbours closer while pushing dissimilar ones apart. Assume there exists an unlabeled dataset \( \mathcal{X} \). For each sentence \( x \in \mathcal{X} \), SimCSE processes the same input through an encoder, such as BERT \cite{kenton2019bert} or RoBERTa \cite{liu2019roberta}, twice. It yields two embeddings \( \boldsymbol{h}_i \) and \( \boldsymbol{h}_i^+ \) for the i-th sentence with different dropout masks. The objective for the pair \( (\boldsymbol{h}_i, \boldsymbol{h}_i^+) \) within a mini-batch of \( B \) is:
\begin{equation}
\mathcal{L}_i = 
-\log \frac{e^{\operatorname{sim}\left(\boldsymbol{h_i}, \boldsymbol{h}_i^{+}\right) / \tau}}{\sum_{j=1}^B e^{\operatorname{sim}\left(\boldsymbol{h}_i, \boldsymbol{h}_j^{+}\right) / \tau}},
\end{equation}
where $\tau$ is a temperature hyperparameter and $\operatorname{sim}(\cdot, \cdot)$ is the cosine similarity between two embeddings. Follow-up methods such as CARDS \cite{wang2022improving}, DiffCSE \cite{chuang2022diffcse}, and RankCSE \cite{liu2023rankcse} have been proposed.

\noindent
\textbf{Data Generation with LLM}. Unsupervised approaches often lag behind their supervised counterparts, which leverage labeled datasets such as natural language inference (NLI) corpora \cite{bowman2015large,williams2018broad}. The NLI dataset provides each \( x \) with a positive sample \( x^+ \) and a hard negative sample \( x^- \) to construct the triplet \( (x, x^+, x^-) \) for the supervised contrastive loss.
However, such annotated data are typically unavailable in most scenarios. Recently, the use of large language models (LLMs) for data construction and annotation has become a trend \cite{li2024optimizing,liu2024socraticlm}. Thus, researchers began exploring the potential of LLMs for the triplet \( (x, x^+, x^-) \) generation for each \( x \in \mathcal{X}\). A representative work is SynCSE \cite{zhang2023contrastive}, which leverages ChatGPT \cite{chatgpt} in a few-shot setting to generate positive samples and hard negative samples. MultiCSR \cite{wang2024large} and GCSE \cite{lai2024enhancing} further refined the process of utilizing LLMs for data generation. These works fundamentally revolve around generating triplets.


\noindent
\textbf{Ranking Sentences Generation}. We further advance this research by concentrating on the generation of ranking sentences. Formally, a ranking sentence is defined as a sequence of sentences  $l=\{x^{(1)},x^{(2)},\cdots,x^{(n)}\}$ for each $x \in \mathcal{X}$, \( x^{(1)} \) is equal to \( x \) itself. Let \( \varphi(a, b) \) denote the semantic similarity between two sentences \( a \) and \( b \), where a larger value indicates a closer semantic similarity. For any three sentences \( (x^{(a)}, x^{(b)}, x^{(c)}) \in l \) with \( a < b < c \), the condition should hold: $\varphi(x^{(a)},x^{(b)}) > \varphi(x^{(a)},x^{(c)})$. In other words, these sentences are arranged in order within the semantic space. 


\section{Methodology}




\subsection{Ranking Sentences Generation}

A straightforward approach to generating ranking sentences is prompting LLM, either by directly producing ranking sentences at once or by generating them step by step. Taking the second case as an example, let the prompt be denoted as an instruction \( I \). The LLM $C_{\theta}$ generates the i-th sentence \( x^{(i)} = [x^{(i)}_1, x^{(i)}_2, \dots, x^{(i)}_k] \) in the following form:
\begin{equation}
p_{\theta}(x^{(i)}|x^{(i-1)},I) = \prod_{t=1}^n p_{\theta}(x^{(i)}_t | x^{(i)}_{<t}, x^{(i-1)}, I),
\label{eq:llm}
\end{equation}
where \( x_{<t}^{(i)} \) represents the tokens generated before the \( t \)-th step. Eq.(\ref{eq:llm}) use the previously generated sentence $x^{(i-1)}$ and the instruction \( I \) to prompt the LLM to generate the next sentence $x^{(i)}$, thereby progressively constructing a sequence of ranking sentences. However, as mentioned before, this method of generation leads to ambiguous semantic relationships among the ranking sentences, as illustrated in Figure~\ref{fig:heat_map} (c). 





\begin{figure}[t]
	\centering
	\includegraphics[width=0.5\textwidth]{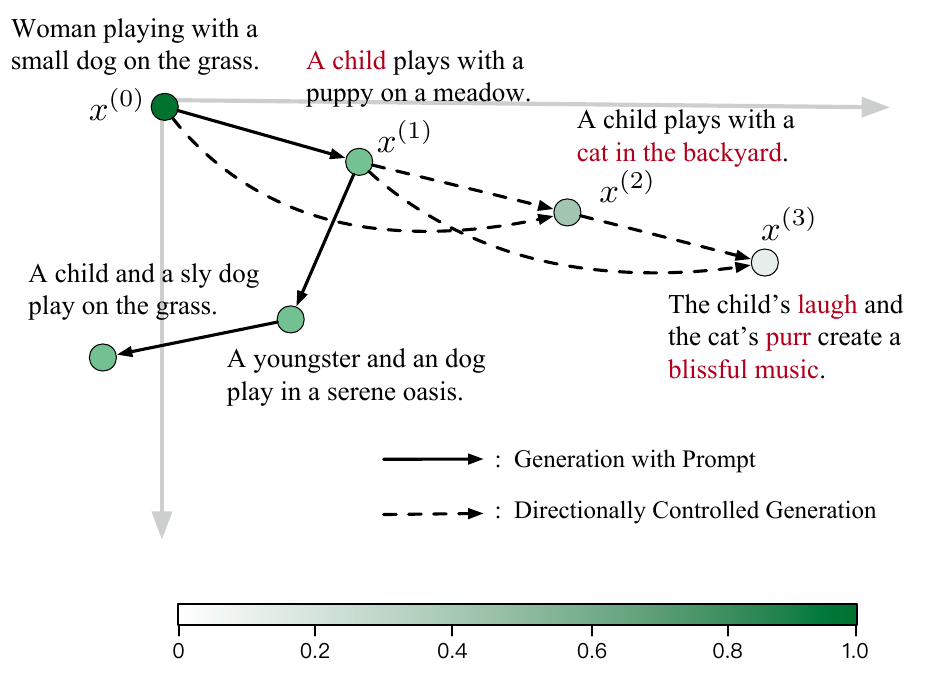}
	\caption{A 2D semantic space illustrating the generation of ranking sentences using prompts (solid line) and our directionally controlled method (dashed line). The color of each point represents its semantic similarity to the initial point \( x^{(0)} \). }
\label{fig:framework}
\end{figure}

Our core idea is to integrate directional control into the ranking sentence generation process. As illustrated in Figure~\ref{fig:framework}, our generation process sequentially combines the directional tendencies of two sentences, ensuring that the subsequent generation maintains a consistent trajectory. For example, the generation direction of \( x^{(3)} \) is controlled by the latent generation space of \( x^{(1)} \) and \( x^{(2)} \), ensuring maximal consistency in their generated directions within the semantic space. Specifically, we modified the sampling method of \( x_t^{(i)} \) in the LLM as follows:
\begin{equation}
\begin{aligned}
p_{\theta}(x_t^{(i)} |x^{(i)}_{<t} , c) \propto \frac{p_{\theta}(x_t^{(i)} |x^{(i)}_{<t} , x^{(i-1)}, I)^{1+\lambda}}{p_{\theta}(x_t^{(i)} |x^{(i)}_{<t} , x^{(i-2)}, I)^{\lambda}}
\end{aligned}
\end{equation}
where \( x^{(i)}_{<t} \) represents the tokens generated before the \( t \)-th step and \( c \) represents the generation condition based on \( x^{(i-1)} \), \( x^{(i-2)} \), and the instruction \( I \). \( \lambda \) is a hyperparameter that assigns weights to the two generation probabilities. 
In other words, the generation of a new sentence depends on the directional tendencies of the generation probabilities of the preceding two sentences, ensuring that the resulting latent space remains aligned in a consistent direction. 
Then, we can thus sample the next t-th token $x_t^{(i)}$ in the logits space:
\begin{equation}
\begin{split}
\log p_{\theta}(x_t^{(i)} &|x^{(i)}_{<t} , c) 
=\\ &(1+\lambda)\log p_{\theta}(x_t^{(i)} |x^{(i)}_{<t} , x^{(i-1)}, I) \\
& - \lambda\log p_{\theta}(x_t^{(i)} |x^{(i)}_{<t} , x^{(i-2)}, I).
\label{eq:prob_sample}
\end{split}
\end{equation}
According to Eq.(\ref{eq:prob_sample}), we concatenate the instruction \( I \) and the previously generated segment \( x^{(i)}_{<t} \) with \( x^{(i-1)} \) and \( x^{(i-2)} \) separately. We then perform two decoding procedures to obtain their respective log probabilities. After computing a weighted sum of these log probabilities, we apply greedy sampling to generate \( x_t^{(i)} \). When generating the first sentence \( x^{(1)} \), since only \( x^{(0)} \) is available, we set \( \lambda \) to 0.

Our method generalizes to Eq.(\ref{eq:llm}) when we set \(\lambda = 0\). However, when we use two sentences as conditions, the generative process undergoes a fundamental transformation. This can be likened to basic geometric theorems: ``Infinitely many lines pass through a single point'' and ``The uniqueness of a line through two points.'' The presence of two preceding sentences ensures the directional consistency of our generation. Besides, our controlled generation process is formally similar to classifier-free guidance \cite{ho2021classifier}, which employs a linear combination to integrate conditional and unconditional score estimations. However, our method differs fundamentally. We rely solely on conditional control, meaning that all terms depend on preceding sentences rather than an unconditional distribution. By doing so, we effectively guide the sentence generation process, ensuring that the generated text maintains a stable and coherent flow within the semantic space.

\subsection{Model Post-training}

After obtaining the ranking sentences $l$ for each \( x \in \mathcal{X} \), we aim to post-train the existing sentence embedding model to enhance its ability to distinguish fine-grained semantic differences. The ranking sentence \( l \) provides order information among sentences. However, the semantic gaps between these sentences are not evenly spaced. Thus, we propose a post-training method that considers both ranking and semantic information among the ranking sentences. 

Let \(\varphi_{j,k} = \varphi(x^{(j)}, x^{(k)})\) denote the semantic similarity between \(x^{(j)}\) and \(x^{(k)}\). For any \( x^{(j)} \in l \), the following semantic ranking relationship should be satisfied according to the ordering within the ranking sentences:
\begin{equation}
\varphi_{j,1} < \cdots < \varphi_{j,j} > \varphi_{j,j+1} > \cdots > \varphi_{j,n}.
\end{equation}
Let \( r = \{ r(i) \}_{i=0}^{n} \) denote a permutation of the object indices arranged in descending order of semantic similarity, where \( r(i) \) represents the rank of the \( i \)-th index in the list \( [\varphi_{j,1}, \varphi_{j,2}, \dots, \varphi_{j,n}] \) based on its magnitude. Then, we process each \( x \) in \( l \) using an encoder model, specifically adopting the DiffCSE \cite{chuang2022diffcse} base series in our experiments. This model can be trained through a standard unsupervised contrastive learning approach. We obtain their corresponding embeddings \( \{ \boldsymbol{h}^{(1)}, \boldsymbol{h}^{(2)}, \dots, \boldsymbol{h}^{(n)} \} \). Assuming \( \phi_{j,k} = \phi(\boldsymbol{h}^{(j)}, \boldsymbol{h}^{(k)}) \) represents the cosine similarity between \( \boldsymbol{h}^{(j)} \) and \( \boldsymbol{h}^{(k)} \).
For \( \boldsymbol{h}^{(j)} \), we can then derive its similarity relationships with other sentences in $l$, represented as  \( \phi^j = [\phi_{j,1}, \phi_{j,2}, \dots, \phi_{j,n}] \). 


Next, we integrate the ranking information \( r \) with the semantic information $\phi^j$. Our fundamental idea is to adjust \( \phi_{j,k} \) based on the ranking position \( r(k) \). Let \( \phi^j[i] \) represent the value at index \( i \) in \( \phi^j \), and let \( \hat{r} = \{ \hat{r}(i) \}_{i=0}^{n} \) denote the permutation of object indices based on the similarity relationships in \( \phi^j \). We modify each \( \phi_{j,k} \) using the following approach:
\begin{equation}
    \hat{\phi}_{j,k}= \left\{
\begin{array}{rcl}
\phi_{j,k} + m(j,k) & \text{if} & r(k) < \hat{r}(k), \\
\phi_{j,k} - m(j,k) & \text{if} & r(k) > \hat{r}(k). \\
\end{array} \right.
\label{eq:alpha}
\end{equation}
\begin{equation}
    m(j,k) = \log{(\omega \cdot |\phi_{j,k}-\phi^j[r(k)] |+1)},
\label{eq:cal_m}
\end{equation}
where $\omega$ is a hyperparameter to control the importance of ranking information. When the ranking order \(\hat{r}\) reflected by semantic information differs from the ranking order \(r\) in the ranking information, the value of \(\phi_{j,k}\) is adjusted based on the ranking discrepancy to bring \(\hat{r}\) closer to \(r\). Through Eq.(\ref{eq:alpha}), we obtain a score \( \hat{\phi}^j = [\hat{\phi}_{j,1}, \hat{\phi}_{j,2}, \dots, \hat{\phi}_{j,n}] \) that seamlessly integrates both ranking information and semantic information. Appendix~\ref{apd:alg} presents the detailed algorithm.



Finally, we post-train the sentence embedding model using the ListMLE \cite{xia2008listwise} loss. Suppose the representation of \( x^{(j)} \) obtained through the sentence embedding model is \( \boldsymbol{e}^{(j)} \). Similarly, we can get a similarity relationships list \( s^j = [s_{j,1}, s_{j,2}, \dots, s_{j,n}] \). The objective of ListMLE for ranking sentence $l$ is defined as:
\begin{equation}
\mathcal{L}_{\text {ListMLE }}(l)=-\sum_{j=1}^n \log P\left(\hat{\phi}^j \mid s^j\right).
\end{equation}
This target ensures that the ranking results produced by the sentence embedding model learn to align with the ranking results obtained from the fusion of ranking information and semantic information in the ranking sentences.





\section{Experiments}

\subsection{Dataset}
Similar to SynCSE \cite{zhang2023contrastive} and MultiCSR \cite{wang2024large}, we utilize the premises of the NLI dataset \cite{bowman2015large,williams2018broad} as the initial unlabeled dataset, denoted as \( X_1 \). Unlike SynCSE and MultiCSR, which employ the full dataset, we sample only a subset for a generation. Specifically, to enhance data diversity, we first apply k-means clustering to \( X_1 \). We set the number of cluster centers to 1,000 and then performed random sampling, selecting 20 samples per cluster, resulting in the dataset \( X_2 \). Next, we generate ranking sentences for each sentence in \( X_2 \) using our method, where the generation step is set to 32, and the hyperparameter \( \gamma \) is set to 1.5. In contrast to SynCSE, which relies on ChatGPT with approximately 175B parameters for generation, we utilize the LLaMA3-8B-Instruct. This process ultimately produces the dataset \( X_3 \), consisting of 16,063 sentence ranking lists and a total of 530,079 sentences. Appendix~\ref{apd:generation} presents the detailed generation process of our method.

\subsection{Experiment Setup}
\noindent
\textbf{Baselines}.  We chose the following strong baselines, including SimCSE \cite{gao2021simcse}, DiffCSE \cite{chuang2022diffcse}, PromptBERT \cite{jiang2022promptbert}, PCL \cite{wu2022pcl}, DebCSE \cite{miao2023debcse}, InfoCSE \cite{wu2022infocse}, RankCSE \cite{liu2023rankcse}, SynCSE \cite{zhang2023contrastive}, and MultiCSR \cite{wang2024large}. 
Our model is built upon existing sentence embedding models as a post-training approach. In the following experiments, we primarily selected two SOTA models, SynCSE and MultiCSR, as our base models to evaluate whether integrating our ranked sentence data and post-training method can enhance performance\footnote{By the time this work was completed, GCSE \cite{lai2024enhancing} was one of the most recent approaches utilizing synthetic data for sentence embedding training. However, as it has not yet publicly released its code and dataset, we did not consider it as a base model.}. We designate the post-trained SynCSE and MultiCSR as SynCSE-r and MultiCSR-r, respectively. The details of our training process are provided in Appendix~\ref{apd:training_details}.


\noindent
\textbf{Evaluation Settings}. We conduct evaluation tests across three tasks: Semantic Textual Similarity (STS), Reranking Task, and Transfer Task (TR).  Specifically, for the STS task, we assess performance on seven STS benchmarks: STS 2012–2016 \cite{agirre2012semeval,agirre2013sem,agirre2014semeval,agirre2015semeval,agirre2016semeval}, STS Benchmark \cite{cer2017semeval}, and SICK-Relatedness \cite{MarelliMBBBZ14}. These datasets consist of sentence pairs annotated with similarity scores ranging from 0 to 5. For the retrieval task, we conduct experiments on four datasets: AskUbuntuDupQuestions \cite{barzilay2016semi}, MindSmallReranking \cite{wu2020mind}, SciDocsRR \cite{wu2020mind}, and StackOverflowDupQuestions \cite{liu2018linkso}. We followed the validation approach of SynCSE \cite{zhang2023contrastive}, adopting the methodology of MTEB \cite{muennighoff2023mteb} and employing Mean Average Precision (MAP) as the primary evaluation metric. For the TR task, we use SentEval \cite{conneau2018senteval} to evaluate the results, as detailed in Appendix~\ref{apd:transfer}.


\subsection{Main Results}

\textbf{STS Tasks.} As shown in Table~\ref{tab:sts_task}, the post-trained model obtained through our method significantly outperforms previous baselines. Compared to the standard unsupervised SimCSE, our SOTA results improve Spearman’s correlation by an average of 7.11\% on base models and 5.09\% on large models. In comparison with ranking-aware models such as RankCSE, our method achieves improvements of 2.19\% and 2.65\%, respectively. Furthermore, compared to the underlying sentence embedding models we employ, such as MultiCSR and SynCSE, our approach enhances performance by 0.61\% and 0.45\% on base models and large models, respectively. These results demonstrate that our method has successfully achieved new SOTA models.

\begin{table*}[!t]
\resizebox{\textwidth}{!}{%
\centering
\begin{tabular}{c|c|cccccccc}
\toprule
\textbf{Model}                 & \textbf{Method}      & \textbf{STS-12} & \textbf{STS-13} & \textbf{STS-14} & \textbf{STS-15} & \textbf{STS-16} & \textbf{STS-B} & \textbf{SICK-R} & \textbf{Avg.}  \\ \midrule
\multirow{10}{*}{BERT-base} 
                               & SimCSE\dag               & 68.40           & 82.41           & 74.38           & 80.91           & 78.56           & 76.85          & 72.23           & 76.25          \\
                               & DiffCSE\dag              & 72.28           & 84.43           & 76.47           & 83.90           & 80.54           & 80.59          & 71.23           & 78.49          \\
                               & PromptBERT$\clubsuit$           & 71.56           & 84.58           & 76.98           & 84.47           & 80.60           & 81.60          & 69.87           & 78.54          \\
                               & PCL$\spadesuit$              & 72.84           & 83.81  & 76.52           & 83.06           & 79.32           & 80.01          & 73.38           & 78.42          \\
                               & DebCSE\dag               & {\ul 76.15}     & {\ul 84.67}           & \textbf{78.91}           & \textbf{85.41} & 80.55 & {\ul 82.99}          & 73.60           & 80.33          \\
                               & InfoCSE$\dag\dag$ & 70.53 & 84.59 & 76.40  &  85.10 & {\ul 81.95}  & 82.00 &  71.37 & 78.85 \\
                               & RankCSE$\spadesuit$              & 75.66           & \textbf{86.27}  & 77.81           & 84.74           & 81.10           & 81.80          & 75.13           & 80.36          \\
                               & SynCSE* & 74.53  & 82.14           & 78.22           & 83.46  & 80.66     & 81.42          & 80.51     & 80.13          \\
                               & MultiCSR* & 75.88           & 82.39           & {\ul 78.80}     & 84.42           & 80.54           & 82.23 & 80.03  & 80.61  \\
                               & SynCSE-r (ours) &  75.82 & 83.24 & 78.61 & 84.75 & 81.68 & 83.45 &  80.67 & {\ul 81.17} \\
                               & MultiCSR-r (ours) & \textbf{76.37}  & 82.50  & 78.37  & {\ul 85.38} & \textbf{82.15}  & \textbf{84.01}  & {\ul 80.55}  & \textbf{81.33} \\
                               
                               \midrule
\multirow{9}{*}{BERT-large}    & SimCSE\dag               & 70.88           & 84.16           & 76.43           & 84.50           & 79.76           & 79.26          & 73.88           & 78.41          \\
                               & PCL$\spadesuit$              & 74.87           & 86.11           & 78.29           & 85.65           & 80.52     & 81.62          & 73.94           & 80.14          \\                               
                               & DebCSE\dag               & \textbf{76.82}     & {\ul 86.36}     & 79.81    & {\ul 85.80}  & 80.83           & 83.45 & 74.67           & 81.11          \\
                               & InfoCSE$\dag\dag$ & 71.89 & 86.17 & 77.72 & \textbf{86.20} & 81.29 & 83.16 & 74.84 & 80.18 \\
                               & RankCSE$\spadesuit$              & 75.48           & \textbf{86.50}           & 78.60           & 85.45           & {\ul 81.09}     & 81.58          & 75.53           & 80.60          \\
                               & SynCSE*  & 75.23 & 84.28 & 79.41 & 84.89 & 82.09  & 83.48  & 81.79  & 81.60  \\
                               & MultiCSR*  & 75.56  & 85.19  & \textbf{80.14} & 85.91 & 82.40  & 84.19 & 81.65  & 82.15 \\
                               & SynCSE-r (ours)  & {\ul 76.32} &  85.17 & 79.29 & 85.78 &  \textbf{82.76} & 84.76  & \textbf{82.51}  & {\ul 82.37} \\
                               & MultiCSR-r (ours)  & 75.69  & 85.63  & {\ul 79.92}  & {\ul 86.08} & {\ul 82.69} & \textbf{84.88}    & {\ul 82.37}  & \textbf{82.47} \\ \midrule
\multirow{10}{*}{RoBERTa-base}  & SimCSE\dag               & 70.16           & 81.77           & 73.24           & 81.36           & 80.65           & 80.22          & 68.56           & 76.57          \\
                               & DiffCSE\dag              & 70.05           & 83.43           & 75.49           & 82.81           & 82.12           & 82.38          & 71.19           & 78.21          \\
                               & PromptRoBERTa$\clubsuit$           & 73.94           & 84.74           & 77.28           & 84.99           & 81.74           & 81.88          & 69.50           & 79.15          \\
                               & PCL$\spadesuit$              & 71.13           & 82.38  & 75.40           & 83.07           & 81.98     & 81.63          & 69.72           & 77.90          \\
                               & DebCSE\dag               & 74.29           & {\ul 85.54}     & 79.46           & 85.68  & 81.20           & 83.96          & 74.04           & 80.60          \\
                              
                               & RankCSE$\spadesuit$              & 73.20           & \textbf{85.95}  & 77.17           & 84.82           & 82.58    & 83.08          & 71.88           & 79.81          \\
                               & SynCSE* & 76.15 & 84.41 & 79.23 & 84.85 & {\ul 82.87} & 83.95          & \textbf{81.41} & 81.84  \\
                               & MultiCSR* & \textbf{77.03} & 84.72  & {\ul 79.71} & 85.80           & 82.68  & 84.24 & 80.64  & {\ul 82.12} \\
                               & SynCSE-r (ours)  & 76.01 & 83.18 & 79.13 &  85.51 & \textbf{83.03} &  {\ul 84.66} & 80.93 & 81.78 \\
                               & MultiCSR-r (ours)& {\ul 76.79}  & 85.03  & \textbf{80.00} & \textbf{86.05}   & 82.65 & \textbf{84.79} & {\ul 81.14} & \textbf{82.35} \\ \midrule
\multirow{8}{*}{RoBERTa-large} & SimCSE\dag               & 72.86           & 83.99           & 75.62           & 84.77           & 81.80           & 81.98          & 71.26           & 78.90          \\
                               & PCL$\spadesuit$              & 74.08           & 84.36           & 76.42           & 85.49           & 81.76           & 82.79          & 71.51           & 79.49          \\                               
                               & DebCSE\dag               & \textbf{77.68}     & \textbf{87.17}     & \textbf{80.53}     & \textbf{85.90}           & {\ul 83.57}     & {\ul 85.36} & 73.89           & 82.01          \\
                               & RankCSE$\spadesuit$              & 73.20           & {\ul 85.83} & 78.00           & 85.63           & 82.67           & 84.19          & 73.64           & 80.45          \\
                               & SynCSE* & {\ul 75.92}  & 85.01           & {\ul 80.43}        & 85.83     & {\ul 84.40}  & 85.05 & {\ul 81.99}  & {\ul 82.66}  \\
                               & MultiCSR* & 74.42 & 84.46 & 79.17 & {\ul 84.76} & 83.67 &  84.23 & 81.50 & 81.74 \\
                               & SynCSE-r (ours) & 75.64  & 84.53 & 80.36  & {\ul 85.88}  & \textbf{84.47} & \textbf{85.82}  & \textbf{83.24}  & \textbf{82.85} \\
                               & MultiCSR-r (ours)&  74.28 & 84.81  & 79.20  &  85.26 &  83.93 & 84.40 & 81.62 & 81.93 \\ \bottomrule
\end{tabular}%
}
\caption{Comparison of Spearman's correlation results on STS tasks, where the value highlighted in bold is the best value, and the value underlined is the second-best value. ``\dag'': results from \cite{miao2023debcse}, ``$\clubsuit$'': results from \cite{wang2024large}, ``$\spadesuit$'': results from \cite{liu2023rankcse}, ``$\dag\dag$'': results from \cite{wu2022infocse}. ``*'': we reproduce the results with the officially released codes and corpus from \cite{zhang2023contrastive,wang2024large}.}
\label{tab:sts_task}
\end{table*}

\noindent
\textbf{Reranking Tasks.} Table~\ref{tab:reranking} presents the performance on four reranking datasets. We followed the experimental setup of SynCSE \cite{zhang2023contrastive} without utilizing the training sets of reranking tasks. During model training, only the synthesized data was used. We compared the changes in MAP for SynCSE and MultiCSR after post-training with our ranking sentences. Overall, our synthesized data and method led to an average improvement of 1.35\% and 1.11\% for SynCSE and MultiCSR, respectively. Note that both SynCSE and MultiCSR employ contrastive learning, which is originally a training paradigm for retrieval models \cite{izacard2021unsupervised,li2021more}. Our synthesized ranking sentences further enhance the reranking performance of SynCSE and MultiCSR, demonstrating their effectiveness in this context.

\begin{table}[!t]
\setlength{\tabcolsep}{3pt} 
\scalebox{0.77}{
\centering
\begin{tabular}{@{}ccccc@{}}
\toprule
\textbf{Dataset} & \textbf{SynCSE} & \textbf{SynCSE-r} & \textbf{MultiCSR} & \textbf{MultiCSR-r} \\ 
\midrule
\multicolumn{5}{c}{BERT-base} \\ 
\hline
AskU. & 51.79 & 52.34 (+1.05\%) & 51.04 & 51.51 (+0.92\%) \\
Mind. & 28.96 & 29.01 (+0.17\%) & 29.04 & 29.37 (+1.14\%) \\
SciD. & 69.49 & 70.73 (+1.79\%) & 69.32 & 70.61 (+1.87\%) \\
StackO. & 39.88 & 40.66 (+1.94\%) & 39.50 & 40.68 (+2.97\%) \\
Avg. & 47.53 & 48.19 (+1.37\%) & 47.22 & 48.04 (+1.73\%) \\
\midrule
\multicolumn{5}{c}{BERT-large} \\ 
\hline
AskU. & 51.36 & 50.73 (-1.22\%) & 51.62 & 50.49 (-2.19\%) \\
Mind. & 30.56 & 30.62 (+0.18\%) & 29.47 & 30.68 (+4.11\%) \\
SciD. & 71.33 & 72.22 (+1.25\%) & 71.31 & 71.71 (+0.56\%) \\
StackO. & 40.06 & 39.82 (-0.60\%) & 39.76 & 40.09 (+0.84\%) \\
Avg. & 48.33 & 48.35 (+0.04\%) & 48.04 & 48.24 (+0.43\%) \\
\midrule
\multicolumn{5}{c}{RoBERTa-base} \\ 
\hline
AskU. & 52.59 & 53.26 (+1.28\%) & 51.91 & 52.18 (+0.52\%) \\
Mind. & 27.58 & 28.70 (+4.06\%) & 27.97 & 28.37 (+1.45\%) \\
SciD. & 63.39 & 65.94 (+4.02\%) & 62.83 & 64.18 (+2.15\%) \\
StackO. & 38.81 & 38.84 (+0.07\%) & 39.35 & 39.95 (+1.53\%) \\
Avg. & 45.59 & 46.69 (+2.39\%) & 45.51 & 46.17 (+1.45\%) \\
\midrule
\multicolumn{5}{c}{RoBERTa-large} \\ 
\hline
AskU. & 55.22 & 54.92 (-0.54\%) & 54.01 & 54.66 (+1.21\%) \\
Mind. & 29.88 & 30.17 (+0.99\%) & 29.16 & 29.32 (+0.56\%) \\
SciD. & 69.33 & 70.99 (+2.39\%) & 69.73 & 70.08 (+0.49\%) \\
StackO. & 39.00 & 40.42 (+3.65\%) & 40.50 & 40.92 (+1.04\%) \\
Avg. & 48.36 & 49.13 (+1.59\%) & 48.35 & 48.75 (+0.82\%) \\
\bottomrule
\end{tabular}}
\caption{Comparison of Mean Average Precision (MAP) results on reranking tasks, illustrating the changes in SynCSE and MultiCSR before and after training with ranking sentence data.}
\label{tab:reranking}
\end{table}

\subsection{Ablation Study}
Since our method consists of both a data generation phase and a model post-training phase, we conduct two groups of ablation experiments. For data synthesis, we design the following three ablation settings: (a) Prompting the LLM to generate
complete ranking sentences at once. (b) Prompting the LLM to generate ranking sentences step by step. (c) Using our method, we first generate ranking sentences. Then, we randomly shuffle them and reconstruct new ranking sentences. In this case, only semantic information is utilized since the ranking information is lost. For the post-training phase, we designed the following two ablation settings: (d) Only ranking information $r$ is used. (e) Only semantic similarity information $\phi^{(j)}$ is used.  

Table~\ref{tab:ablation_study} presents the average Spearman's correlation on the STS dataset. For data synthesis, comparing (a) shows that generating ranking sentences at once via prompts is limited, as LLMs struggle with semantic understanding in longer texts. Comparison with (b) suggests that even a step-by-step approach lacks effective directional control, leading to suboptimal results. The results of (c) highlight the importance of ranking information, confirming that our method’s improvements are not solely due to semantic information. For post-training, comparing (d) indicates that ranking information alone is insufficient due to fine-grained semantic differences, emphasizing the need for semantic information. The results of (e) remain inferior to the full method, showing that incorporating ranking information helps refine semantic representations and improve model performance.


\begin{table}[!t]
\centering
\scalebox{0.8}{
\begin{tabular}{cccc}
\toprule
\textbf{Phase}  & \textbf{Method}  & \textbf{Spearman's} &  $\Delta$ \\ 
\midrule
- & MultiCSR-r & 81.33 & 0.0 \\
Data Generation & (a)  & 80.85 & -0.48 \\
Data Generation & (b)  & 80.97 & -0.36 \\
Data Generation & (c)  & 80.78 & -0.55 \\
Post-training & (d)  & 80.65 & -0.68 \\ 
Post-training & (e)  & 81.18 &  -0.15 \\ 
- & MultiCSR  & 80.61 & -0.72 \\ 
\bottomrule
\end{tabular}}
\caption{Ablation studies on different data generation methods and components of post-training. We use MultiCSR-r based on BERT-base as the model, and the results are reported on average Spearman's correlation of STS Task.}
\label{tab:ablation_study}
\end{table}

\subsection{Analysis}

In this section, we conduct a more in-depth analysis of the synthesized dataset and our post-training method. We employ SynCSR-r and MultiCSR-r with the BERT-base model. We report the results in terms of Spearman’s correlation on the STS task.

\begin{figure}[t]
\centering
\subfigure[impact of $\omega$ ]{
\label{fig:back1}
\includegraphics[width=0.23\textwidth]{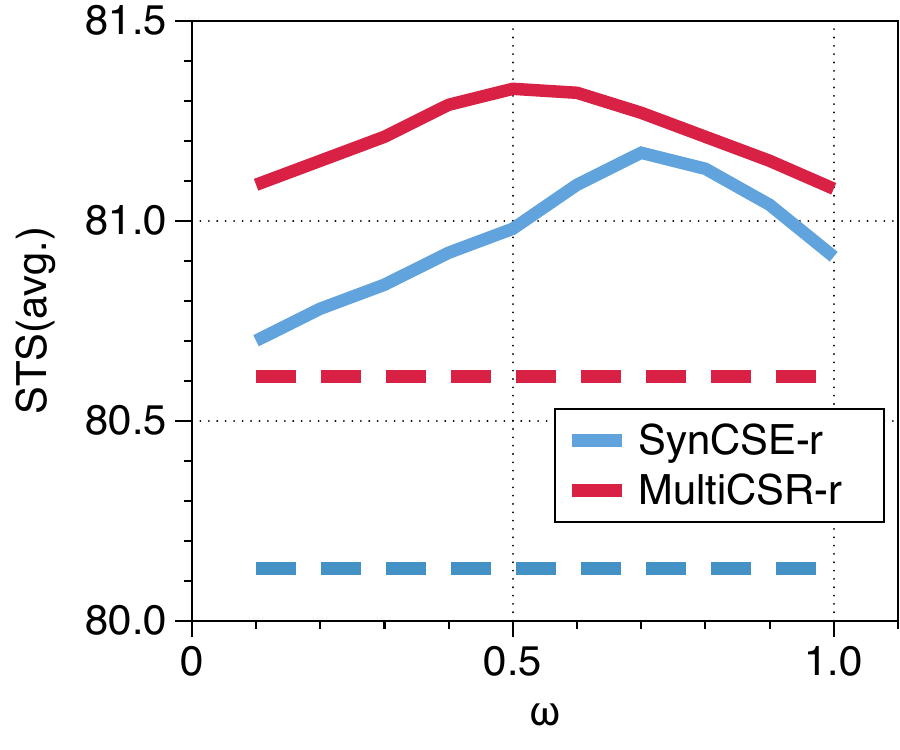}} \hspace{-0.3cm}
\subfigure[impact of data]{
\label{fig:back2}
\includegraphics[width=0.23\textwidth]{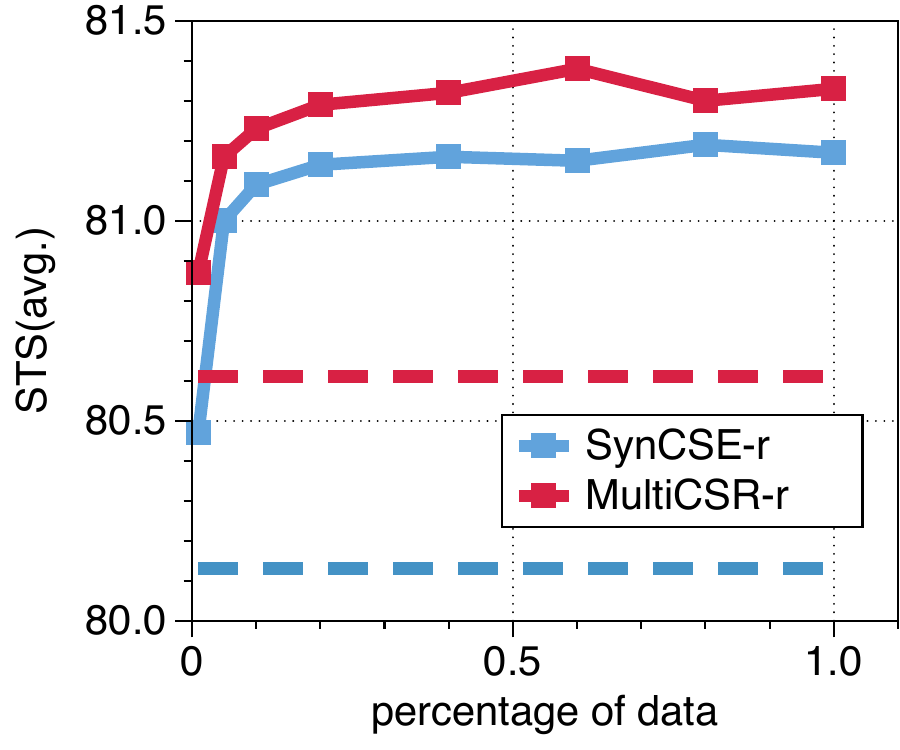}}
\caption{The impact of hyperparameter $\omega$ on average STS test score for SynCSE-r and MultiCSR-r based on BERT-base as the model. The base model scores are shown in dashed lines}
\label{fig:analysis}
\end{figure}

\noindent
\textbf{The impact of the hyperparameter \(\omega\).} Figure~\ref{fig:analysis} (a) illustrates the impact of different hyperparameter \(\omega\) on model performance in the STS task. The \(\omega\) plays a crucial role in our post-training method, as defined in Eq.~(\ref{eq:cal_m}), where it controls the importance of ranking information. The results indicate that while the optimal \(\omega\) varies across different models, it remains robust within a relatively broad range. Based on these findings, we set \(\omega = 0.7\) for SynCSE-r and \(\omega = 0.5\) for MultiCSR-r.


\begin{table*}[!t]
\resizebox{\textwidth}{!}{%
\centering
\begin{tabular}{c|cccccccc}
\toprule
 \textbf{Method}      & \textbf{STS-12$(\Delta)$} & \textbf{STS-13$(\Delta)$} & \textbf{STS-14$(\Delta)$} & \textbf{STS-15$(\Delta)$} & \textbf{STS-16$(\Delta)$} & \textbf{STS-B$(\Delta)$} & \textbf{SICK-R$(\Delta)$} & \textbf{Avg.$(\Delta)$}  \\ \midrule
 SimCSE    & +4.04 & +1.47  & +1.61 & +2.31 & +0.49 & +1.87 & +0.66 & +1.78 \\
                                InfoCSE              &    +0.24        &   -0.42         &    +0.27       &   +0.36    &     +0.64       &  +0.14  &    +1.38     &  +0.37     \\
                                PCL &  +0.69 &  +0.69       &    +0.25       &    +2.49     &   +2.79   & +2.27  & +1.4  & +1.51  \\
                                RankCSE  & -0.11  & +0.6  & +0.69 & +2.00   &  +1.98 & +0.15 & -0.29 &  +0.71  \\
            \bottomrule
\end{tabular}%
}
\caption{We compare the changes in Spearman’s correlation on STS tasks across several sentence embedding models after post-training with ranking sentences. Their checkpoints based on BERT-base as the model are obtained from their official sources.}
\label{tab:postt}
\end{table*}

\noindent
\textbf{The impact of the amount of synthetic data amount.} Figure~\ref{fig:analysis} (b) illustrates the performance on the STS task when using different proportions of our synthesized ranking sentences dataset. We find that although approximately 16,000 sentence ranking lists were generated, utilizing only \(10\%\) of the data is sufficient to achieve a substantial improvement, while merely \(5\%\) is enough to surpass the original model significantly. This underscores the pivotal role of ranking sentences in enhancing sentence embedding models. On the other hand, we also observe that continuously increasing the number of ranking sentences does not lead to a consistent improvement in STS performance. This may be attributed to an excessive number of ranking sentences, potentially reducing the model's generalization ability.



\begin{table}[!t]
\centering
\begin{tabular}{ccc}
\toprule
\textbf{}  & \textbf{Stage 1}  & \textbf{Stage 2}  \\ 
\midrule
\textbf{Setting 1} & 80.91  & 81.63 \\
\textbf{Setting 2} & 77.12  & 80.74  \\
\bottomrule
\end{tabular}
\caption{Multi-stage post-training analysis. We used two different post-training sequences and conducted experiments on the STS task, recording the average Spearman’s correlation for each stage.}
\label{tab:multi_post}
\end{table}

\subsection{Post-training Experiment}
\label{exp:post}
We further analyze the impact of post-training our data on sentence embedding models other than SynCSE and MultiCSR. Table~\ref{tab:postt} presents the changes in Spearman’s correlation for SimCSE \cite{gao2021simcse}, InfoCSE \cite{wu2022infocse}, PCL \cite{wu2022pcl}, and RankCSE \cite{liu2023rankcse} on STS tasks after applying our ranking sentences dataset and post-training approach. From these results, we can observe that employing ranking sentences along with our post-training method has led to improvements across most datasets in the STS task. This demonstrates the versatility and effectiveness of both ranking sentences and our post-training approach. The detailed results are presented in Figure~\ref{tab:postppp} of the Appendix~\ref{apd:post}.

Additionally, we conducted further experiments to examine the impact of multi-stage post-training. Specifically, we utilized the SimCSE-BERT-base model and assessed two different post-training sequences: \textbf{Setting 1}: We first post-trained using the MultiCSR dataset and method, followed by our dataset and method. \textbf{Setting 2}: We began with post-training on our dataset and method, followed by the MultiCSR dataset and method. Table~\ref{tab:multi_post} shows these results. In Setting 1, we see that our method improves model performance after post-training with the MultiCSR dataset, achieving a new SOTA performance of 81.63. This highlights the flexibility and compatibility of our approach with different post-training strategies. Moreover, comparing the two settings, MultiCSR offers a greater performance boost when used as the first stage of post-training. This supports our claim that our ranking sentence dataset and method are especially effective in enhancing fine-grained semantic understanding in sentence embedding models.

\begin{figure}[t]
\centering
\subfigure[average length]{
\label{fig:data_a_1}
\includegraphics[width=0.23\textwidth]{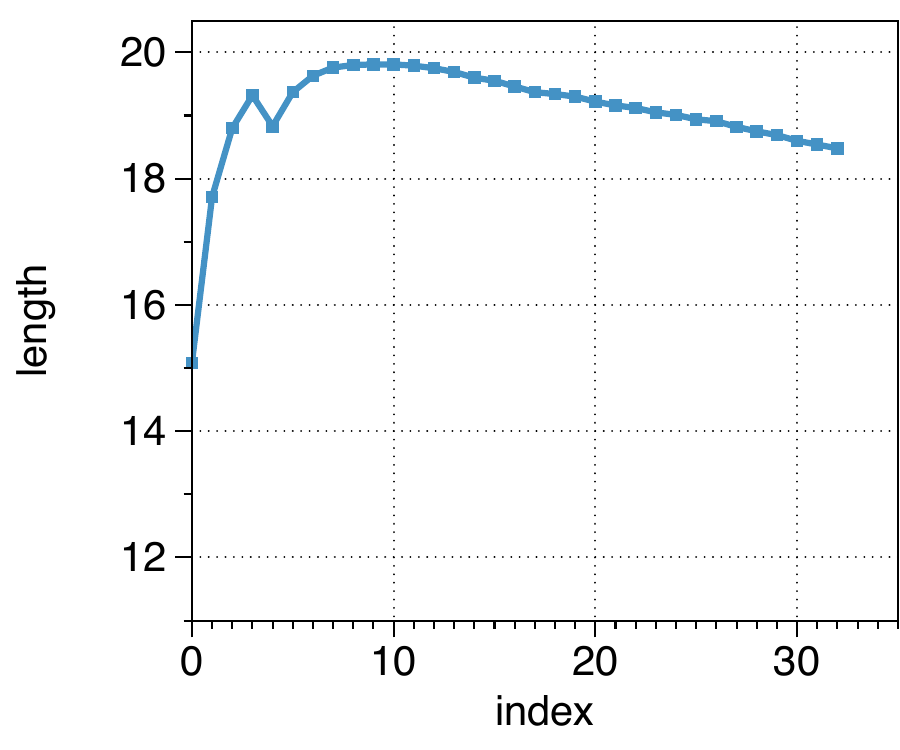}} \hspace{-0.3cm}
\subfigure[sentence quality]{
\label{fig:data_a_2}
\includegraphics[width=0.23\textwidth]{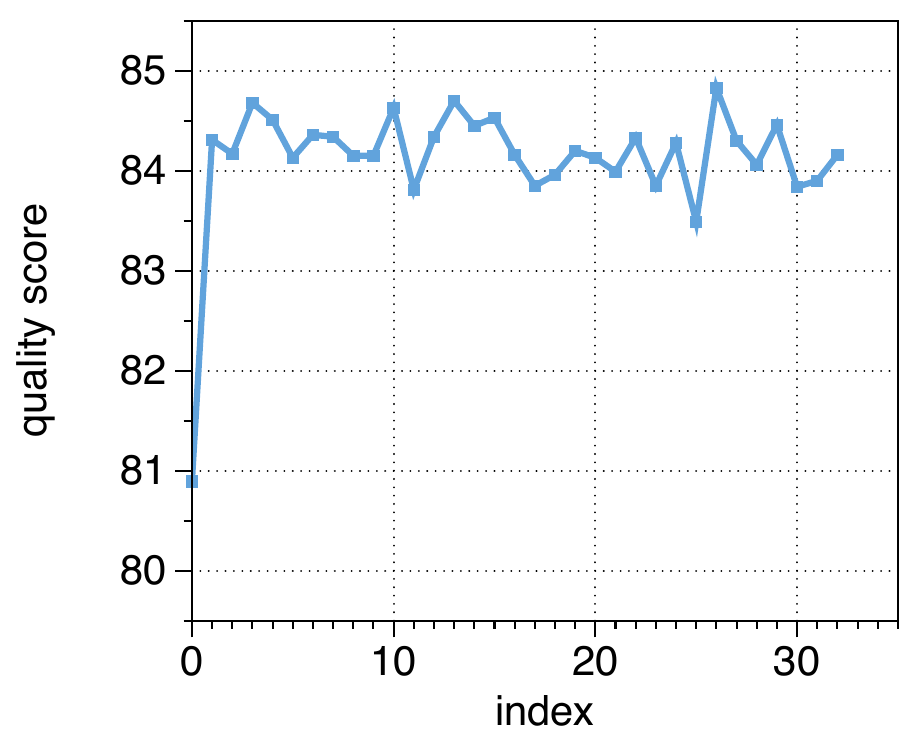}}
\caption{The analysis of the generated ranking sentences involved calculating the average length (in terms of word count) and the average sentence quality for each index. An index of 0 represents the original sentence.}
\label{fig:data_analysis}
\end{figure}

\subsection{Ranking Sentences Analysis}

We further designed two analytical experiments for the synthetic dataset. First, a previous study \cite{an2024capturing} reports that sentences synthesized using large language models (LLMs) tend to be longer than the original text. Therefore, we investigated whether the generated ranking sentences also exhibit this characteristic. We calculated the average length of sentence tokens for each ranking index of the generated sentences, and the results are presented in Figure~\ref{fig:data_a_1}. The findings indicate that our method produces slightly longer sentences than the originals, but this increase stabilizes, preventing excessive redundancy in subsequent sentences.

Second, we analyzed the quality of the generated sentences. Inspired by the Individual Score method \cite{chen2023exploring}, we employed GPT-4o to evaluate each sentence's quality on a scale from 0 (worst) to 100 (best). We randomly selected 200 ranking sentences along with their original data for this evaluation. Figure~\ref{fig:data_a_2} displays the results. Compared to the original sentences, the generated sentences demonstrate higher quality. These results suggest that the quality of the generated sentences can be reliably assured.




\section{Related Work}

Unsupervised sentence embedding has been widely studied. Early methods extended the word2vec framework \cite{mikolov2013distributed} to sentence-level embeddings, such as Skip-Thought \cite{kiros2015skip}, FastSent \cite{hill2016learning}, and Quick-Thought \cite{logeswaran2018efficient}. With the rise of PLMs, models like BERT \cite{kenton2019bert} and RoBERTa \cite{liu2019roberta} have been explored for sentence representation. However, issues like anisotropy \cite{ethayarajh2019contextual} have led to post-processing techniques such as BERT-flow \cite{li2020sentence} and BERT-whitening \cite{su2021whitening} to improve embedding quality.

With the rise of contrastive learning, the focus shifted toward deriving sentence embeddings by maximizing agreement between different views of the same sentence. Techniques like SimCSE \cite{gao2021simcse} utilized dropout-based augmentation to create positive pairs, inspiring follow-up methods \cite{wang2022improving,chuang2022diffcse,liu2023rankcse,jiang2022promptbert,wu2022pcl,miao2023debcse}. These methods proved highly effective. However, unsupervised approaches often lag behind their supervised counterparts, which leverage labeled datasets such as natural language inference (NLI) corpora \cite{bowman2015large,williams2018broad}. Yet, such datasets are not easily accessible due to the high annotation cost.

To address these limitations, researchers began exploring sentence generation for unlabeled data \cite{chen2022generate,ye2022progen} using models like T5 \cite{chung2024scaling}. With the advent of large language models (LLMs), both data annotation and generation have seen significant improvements \cite{gilardi2023chatgpt,alizadeh2023open,wang2024improving,alizadeh2025open}. SynCSE \cite{zhang2023contrastive} leverages LLMs to generate semantically similar sentence pairs, enhancing the effectiveness of contrastive learning. MultiCSR \cite{wang2024large} and GCSE \cite{lai2024enhancing} further refine the utilization of LLMs for data generation. This line of research builds upon the training paradigm of supervised SimCSE \cite{gao2021simcse}, where a triplet is generated for contrastive learning. In contrast to these works, our approach shifts the generation objective towards ranking sentence generation, introducing a novel refinement strategy for contrastive learning models.

\section{Conclusion}

In this paper, we investigate a method for synthesizing ranking sentences by leveraging LLMs to generate sentences progressively increasing semantic divergence, guided by a controlled direction in the latent space. Furthermore, we explore a post-training approach that integrates ranking information and semantic information. Experimental results demonstrate that our method achieves new SOTA performance with minimal cost in ranking sentence synthesis.

\section{Limitations}

Although our work has achieved a new SOTA performance for existing sentence embedding models, several promising directions still need to be explored. In the realm of data synthesis, this paper primarily concentrates on the process of data generation. However, the selection and refinement of synthesized data are also crucial. Besides, during the post-training process, our primary approach is to integrate ranking and semantic information. This process involves a hyperparameter \(\omega\), whose magnitude influences the model's performance. Exploring an adaptive method to eliminate the dependence on \(\omega\) is also a worthwhile consideration.

\section*{Acknowledgments}

This research was partially supported by the National Natural Science Foundation of China (Grants No.U23A20319, 62477044), the Fundamental Research Funds for the Central Universities (No.WK2150110038), and the National Key Laboratory of Human-Machine Hybrid Augmented Intelligence, Xi’an Jiaotong University (No. HMHAI-202410).


\bibliography{custom}

\clearpage

\appendix

\section{Ranking Sentences Generation Details}
\label{apd:generation}

In this section, we present the detailed methodology of several ranking sentence generation methods involved in this paper and the detailed methodology of our method. We employ the premises from the NLI dataset \cite{bowman2015large,williams2018broad} as the initial unlabeled dataset for these methods. 
\begin{enumerate}
\item \textbf{Single-step Generation.} Prompting the LLM to generate ranking sentences at once. Our preliminary experiments reveal that generating complete ranking sentences in a single step is too challenging for the LLaMA3-8B-Instruct model. Therefore, we employ the LLaMA3-70B-Instruct model and adopt a few-shot approach to guide the LLM in the generation, ensuring both the coherence and usability of the generated ranking sentences.
\item \textbf{Iterative Step-by-step Generation.} Prompting the LLM to generate ranking sentences step by step. Specifically, based on the result of the previous sentence, we prompt the LLaMA3-8B-Instruct to generate the next sentence. This process continues until 32 sentences have been generated.
\item \textbf{Our Method.} Generating the ranking sentences using our proposed directionally controlled generation method. We employ the LLaMA3-8B-Instruct model to generate ranking sentences. Specifically, for the first generation, we prompt the LLM to generate sentence $x^{(2)}$. Then, as designed in our method, each input consists of the previous two sentences along with an instruction. By adjusting the sampling strategy of the 8B LLaMA3 model, we progressively generate the final ranking sentences, setting \(\gamma\) to 1.5.
\end{enumerate}

For all generation methods, we employ a rule-based verification process at the final stage to ensure that the generated results are as complete and non-redundant as possible. The first generation method is performed on a Linux server equipped with 8 NVIDIA A800 GPUs, while the second and third methods are conducted on a Linux server with 8 NVIDIA GeForce RTX 4090 GPUs.

\begin{tcolorbox}[title=Prompt for directly generating complete ranking sentences]

Your task is to take an input sentence and generate a sequence of 32 sentences that gradually and progressively diverge in meaning from the original sentence. The final sentence should be completely unrelated to the original sentence.

Example Input:
The cat is sleeping on the warm windowsill.

Example Output:

1. The cat is resting on the cozy windowsill.

2. The cat is lying on a soft cushion by the window.

3. A small animal is curled up near the window.

... [Omit the following sentence list here for conciseness.]

Here is the sentence: \{sentence\}

Each sentence should be similar in length to the original sentence. Do not explain yourself or output anything else.
\end{tcolorbox}

\begin{tcolorbox}[title=Prompt for generating ranking sentences step by step]
Rewrite the following sentence in a way that slightly changes the meaning while keeping it semantically close. The new sentence should not be an exact paraphrase but should introduce a subtle variation in meaning. Do not lose the core idea of the original sentence.

Here is the sentence: \{sentence\}

Your response should be similar in length to the original sentence. Do not explain yourself or output anything else.
\end{tcolorbox}

\begin{tcolorbox}[title=Prompt for our method]
Rewrite the following sentence or phrase using different words and sentence structure while preserving its original meaning. 
Directly answer with the rewritten sentence. Don't give any explanation or description other than the rewritten sentence.

Write a sentence that is entailment with:\{sentence\}. 

Result:
\end{tcolorbox}

\section{Case Study}
\label{apd:case_study}

In this section, we present a case study to illustrate the generated results of our approach, the single-step generation method, and the iterative step-by-step generation method. Table~\ref{tab:case_study} presents the top 10 generated sentences produced by different methods for a given input sentence. We employ the BGE-m3\cite{chen2024bge} model to obtain their embeddings and compute the cosine similarity between the generated results and the original sentence. Similarity scores for results that are not ranked in descending order of semantic similarity are highlighted in red. We observe that, compared to the other two generation methods, our approach produces results that adhere more closely to a descending order in the semantic space. Moreover, as the generation progresses, the likelihood of producing results that deviate from the expected order increases. This underscores the importance of controlling the direction of generation.


\begin{table*}[!t]
\centering
\scalebox{0.75}{
\begin{tabular}{p{6.5cm} p{6.5cm} p{6.5cm}} 
\toprule
\multicolumn{3}{p{19.5cm}}{\centering \textbf{Original Sentence:} A young man wearing a knit cap with the word PARIS on it and a blue jacket on the street.} \\ 
\midrule  
\textbf{Our Method}  & \textbf{Iterative Step-by-step Generation}  & \textbf{Single-step Generation} \\ 
\midrule
A stylish young man, sporting a PARIS-emblazoned knit cap and a blue jacket, strolled down the street. [0.8016]
& A young man, sporting a knit cap with the word PARIS emblazoned on it, walked down the street in a blue jacket. [0.8660]
& A young man wearing a warm hat with a city name on it and a casual jacket outdoors. [0.6829] \\
\midrule
As he walked, the stylish young man's PARIS-emblazoned knit cap and blue jacket made a stylish statement down the street. [0.7996]
& A young man, sporting a knit cap with the word PARIS emblazoned on it, walked down the street in a blue jacket. [0.8342]
&  A person wearing a soft hat with a name tag and a casual coat in public. [0.5749]
\\
\midrule
His outfit, featuring a PARIS-emblazoned knit cap and blue jacket, turned heads as he strolled down the street. [0.7367]
& A young man, sporting a knit cap with a bold, cursive "PARIS" emblazoned on it, strolled down the street in a navy blue jacket. \textcolor{red}{[0.8016]}
& Someone wearing a hat with a label and a casual coat on the sidewalk. [0.5525]\\
\midrule
As he walked down the street, people couldn't help but notice his outfit, which boasted a knit cap adorned with the iconic PARIS logo and a striking blue jacket. [0.7313]
& A young man, sporting a knit cap adorned with a bold, cursive "PARIS" script, walked down the street in a navy blue coat. [0.8077]
& A person wearing a hat with a logo and a casual coat on the pavement. \textcolor{red}{[0.5861]} \\
\midrule
Noticing his outfit became second nature to passersby as he strolled down the street, the knit cap's PARIS logo and striking blue jacket standing out in bold relief. [0.7246]
& A young man, sporting a knit cap bearing a bold, cursive "PARIS" script, strolled down the street in a navy blue overcoat. [0.7801]
& A man wearing a hat with a brand name and a casual coat in the city. \textcolor{red}{[0.6443]} \\
\midrule
As he walked down the street, it soon became automatic for people to notice his outfit, with the PARIS logo on his knit cap and the striking blue jacket making a bold statement. [0.7064]
& A young man, sporting a knit cap adorned with a striking, cursive "PARIS" motif, ambled down the sidewalk in a navy blue overcoat. \textcolor{red}{[0.7460]}
& A man wearing a hat with a logo and a casual coat on the corner. \textcolor{red}{[0.5745]} \\
\midrule
The moment he strolled along the sidewalk, passersby couldn't help but be drawn to his attire, particularly the eye-catching PARIS emblem on his beanie and the vibrant blue coat that turned heads. [0.6063]
& A young man, sporting a knit cap featuring a bold, cursive "PARIS" logo, strolled leisurely along the sidewalk in a navy blue coat. \textcolor{red}{[0.7859]}
& A man wearing a hat with a brand name and a casual coat in the morning. \textcolor{red}{[0.5850]} \\
\midrule
As soon as he ambled down the sidewalk, people walking by were irresistibly captivated by his outfit, specifically the striking PARIS logo on his beanie and the radiant blue coat that commanded attention. 
 \textcolor{red}{[0.5788]}
& A young man, sporting a knit cap adorned with a prominent, cursive "PARIS" logo, ambled along the sidewalk in a navy blue overcoat. [ 0.7683]
& A man wearing a hat with a label and a casual coat in the evening. \textcolor{red}{[0.5787]} \\
\midrule
The moment he strolled down the sidewalk, passersby couldn't help but be drawn in by his ensemble, particularly the eye-catching PARIS logo on his beanie and the dazzling blue coat that demanded notice. 
 [0.5868]
& A young man, sporting a knit cap bearing a large, cursive "PARIS" logo, strolled leisurely along the sidewalk, clad in a navy blue overcoat. \textcolor{red}{[0.7792]}
& A person wearing a hat with a logo and a formal coat at a party. [0.5256]\\
\midrule
As soon as he walked down the street, people couldn't resist being captivated by his outfit, specifically the striking PARIS emblem on his hat and the mesmerizing blue coat that commanded attention. 
 [0.5829]
& A young man, wearing a knit cap adorned with a prominent, cursive "PARIS" emblem, ambled along the sidewalk, wrapped in a navy blue overcoat. [0.7471]
& Someone wearing a hat with a brand name and a formal dress at a wedding. [0.5064] \\
\bottomrule
\end{tabular}}
\caption{A case study is conducted to compare our generation method with the Iterative Step-by-Step Generation and Single-Step Generation approaches. The similarity to the original sentence is indicated at the end of each sentence, highlighted in red if not ranked in descending order of semantic similarity.}
\label{tab:case_study}
\end{table*}


\section{Model Training Details}
\label{apd:training_details}

All our experimental code is implemented using Python and the PyTorch library. The experiments were conducted on a Linux server equipped with eight NVIDIA GeForce RTX 4090 GPUs. We utilized the official implementations for SynCSE \cite{zhang2023contrastive} and MultiCSR \cite{wang2024large}. Specifically, SynCSE provides both model training code and a synthesized dataset. We used their code and dataset to train SynCSE models, including BERT-base, BERT-large, RoBERTa-base, and RoBERTa-large. MultiCSR offers both training and data synthesis code, which we employ to generate data before proceeding with MultiCSR model training.  Additionally, in Section~\ref{exp:post}, we reproduce several models, including SimCSE \cite{gao2021simcse}, InfoCSE \cite{wu2022infocse}, PCL \cite{wu2022pcl}, and RankCSE \cite{liu2023rankcse}. We downloaded their checkpoints from the official HuggingFace repositories and applied our post-training method. During post-training, each model receives a ranking sentence as input per training step. We use the Adam \cite{kingma2014adam} optimizer and set the learning rate to \(3 \times 10^{-6}\). For SynCSE, the hyperparameter \(\omega\) is set to 0.7, while for the other models, \(\omega\) is set to 0.5.

\section{Algorithm to Get $\hat{\phi}^j$}
\label{apd:alg}
We propose an Algorithm~\ref{alg:trans} for efficiently computing \(\hat{\phi}^j\) for \(j = 1,2,\dots,n\). This algorithm takes a similarity relation matrix \(\Phi = [\phi^1, \phi^2, \dots, \phi^n]\) as input and outputs a refined matrix \(\hat{\Phi} = [\hat{\phi}^1, \hat{\phi}^2, \dots, \hat{\phi}^n]\). For each \(\phi^j\), it is unnecessary to compute the full set of values. We only need to calculate the results from index \(i\) onward and then leverage the symmetry of the similarity matrix to complete the remaining entries. The complexity of the algorithm is \(O(n^3 )\). Since \(n\) represents the length of the ranking sentences and is a finite value, the computational complexity of this algorithm is significantly lower than that of the model's inference process.

\begin{algorithm}[t]
    \renewcommand{\algorithmicrequire}{\textbf{Input:}}
    \renewcommand{\algorithmicensure}{\textbf{Output:}}
    \caption{Refined Semantic Similarity Computation}
    \label{alg:trans}
    \begin{algorithmic}[1]
        \REQUIRE Initial similarity matrix \( \Phi \), hyperparameter \( \omega \).
        \ENSURE Refined similarity matrix \( \hat{\Phi} \).
        \STATE Initialize \( A \leftarrow \mathbf{0} \in \mathbb{R}^{n \times n} \).
        \FOR{each row index \( i = 1 \) to \( n \)}
            \STATE Extract the subarray \( \Phi[i, i:n] \).
            \STATE Sort the subarray in descending order and obtain sorted indices.
            \FOR{each column index \( j = i \) to \( n \)}
                \STATE Find the position index $j'$ of \( \Phi[i, j] \) in the sorted array.
                \STATE Compute \( A[i, j] = \Phi[i, j] - \Phi[i,j'] \).
            \ENDFOR
        \ENDFOR
        \STATE Fill \( A \) symmetrically: \( A[j, i] = A[i, j] \) for \( j > i \).

        \STATE Compute \( \hat{\Phi} = \Phi + \text{sign}(A) \cdot \log( \omega \cdot |A| + 1) \).

        \RETURN \( \hat{\Phi} \).
    \end{algorithmic}
\end{algorithm}

\section{Post-training Experiments}
\label{apd:post}

In the preceding experiments, we demonstrated the changes in Spearman’s correlation for SimCSE \cite{gao2021simcse}, InfoCSE \cite{wu2022infocse}, PCL \cite{wu2022pcl}, and RankCSE \cite{liu2023rankcse} on STS tasks before and after training with our data and methodology. In this section, we present the comprehensive results, as illustrated in Figure~\ref{tab:postppp}.

\begin{table*}[!t]
\centering
\resizebox{0.8\textwidth}{!}{%
\begin{tabular}{c|cccccccc}
\toprule
 \textbf{Method}      & \textbf{STS-12} & \textbf{STS-13} & \textbf{STS-14} & \textbf{STS-15} & \textbf{STS-16} & \textbf{STS-B} & \textbf{SICK-R} & \textbf{Avg.}  \\ \midrule
 SimCSE    &   66.05     &    81.49     &   73.61  & 79.73          &     78.12  & 76.52  &  71.86 &  75.34 \\
  SimCSE-r    &   70.09     &     82.96    &    75.22       &    82.04       &   78.61  &  78.39 & 72.52  &  77.12 \\
\midrule
                                InfoCSE              & 70.23           &    84.05        &   75.98        &  84.78     &    81.72        &  81.75  &    71.09     &  78.51     \\
                                InfoCSE-r              &  70.47          &   83.63         &    76.25       & 85.14      &    82.36        &  81.89  &    72.47     &  78.88     \\
                                \midrule
                                PCL           &   73.46    & 81.57        &     74.91      &    82.24     &   79.94   &  79.41 &     71.76      &  77.61 \\
                                PCL-r           &   74.15    &   82.26      &  75.16         &     84.73    &  82.73    &  81.68 &   73.16  & 79.12  \\
                                \midrule
                                RankCSE    & 74.55  & 85.13  &   77.67  &  84.23  &    81.18      &  81.6     & 74.28       & 79.81   \\
                                RankCSE-r            & 74.44  & 85.73  &  78.36  &  86.23 &  83.16  & 81.75 &  73.99 & 80.52   \\
            \bottomrule
\end{tabular}%
}
\caption{We compare Spearman’s correlation on STS tasks across several sentence embedding models after post-training with ranking sentences. Their checkpoints based on BERT-base as the model are obtained from
their official sources.}
\label{tab:postppp}
\end{table*}

\section{Transfer Task}
\label{apd:transfer}

For the TR tasks, we evaluate our method on seven datasets using the default configurations from SentEval: MR \cite{pang2005seeing}, CR \cite{hu2004mining}, SUBJ \cite{pang2004sentimental}, MPQA \cite{wiebe2005annotating}, SST-2 \cite{socher2013recursive}, TREC \cite{voorhees2000building}, and MRPC \cite{dolan2005automatically}. Table~\ref{t5} shows the results. Overall, we have achieved a new SOTA performance on RoBERTa-base and RoBERTa-large. On BERT-base and BERT-large, both SynCSE-r and MultiCSR-r have demonstrated improvements compared to the results after post-training. Furthermore, our enhancement on the MRPC task is particularly significant. This is because MRPC focuses on distinguishing the similarity between sentence pairs, and by incorporating ranking sentences in post-training, the model becomes more adept at capturing fine-grained semantic differences.

\begin{table*}[h!]
\centering
\resizebox{0.9\textwidth}{!}{%
\begin{tabular}{c|c|cccccccc}
\toprule
\textbf{Model}                          & \textbf{Method}              & \textbf{MR}             & \textbf{CR}             & \textbf{SUBJ}           & \textbf{MPQA}           & \textbf{SST2}           & \textbf{TREC}           & \textbf{MRPC}           & \textbf{Avg.}           \\ \midrule
\multirow{7}{*}{BERT-base}     & SimCSE$\spadesuit$              &  81.18         & 86.46          & 94.45          & 88.88          & 85.50    & {\ul 89.80}       & 74.43         & 85.81   \\
                               & DiffCSE$\spadesuit$             & {\ul 81.76}     & 86.20        & {\ul 94.76}   & 89.21    & 86.00    & 87.60          &75.54    & 85.80      \\
                               & PromptBERT$\clubsuit$             &     80.74     &  85.49  &  93.65   &  89.32      &     84.95      &    88.20  &    76.06     &     85.49    \\
                               & PCL$\spadesuit$                 & 80.11      & 85.25         & 94.22      & 89.15         & 85.12          & 87.40       & 76.12     & 85.34     \\
                               & RankCSE$\spadesuit$             & \bf 83.07     & {\ul 88.27}       & \bf 95.06     & {\ul 89.90}      & \bf 87.70      & \bf 89.40       & {\ul 76.23}     & \bf 87.09      \\
                               & SynCSE*  & 81.09 & \bf 88.29 & 93.53 &   \bf 90.02    & 86.60        &  84.40  &   75.30    & 85.60    \\
                               & MultiCSR*  & 81.64   & 87.79  & 93.83 &    89.91   &   87.15      &  80.20  & 75.25      &   85.11  \\
                               & SynCSE-r  &  81.13  & 87.82 & 94.07 &  89.87  &    {\ul 87.42}     &  83.80  & \bf 77.86 & {\ul 86.00}   \\
                               & MultiCSR-r  &  81.47  & 87.53 & 93.99 &   89.68    &  86.55       &  83.80  &   76.00    &  85.57   \\
                               \midrule
\multirow{5}{*}{BERT-large}    & SimCSE$\spadesuit$  & \bf 85.36       & 89.38        & {\ul 95.39} & 89.63     &  90.44  & 91.80       & 76.41       & {\ul 88.34}        \\
                               & PCL$\spadesuit$   & 82.47  & 87.87      & 95.04         & 89.59       & 87.75      & {\ul 93.00}        & 76.00       &   87.39    \\
                               & RankCSE$\spadesuit$    & 84.63 & 89.51    & \bf 95.50          & \bf 90.08      & \bf 90.61     & \bf 93.20        & {\ul 76.99}  & \bf 88.65     \\
                               & SynCSE*  &  84.66  & 89.96 & 94.49 &  \bf 90.08  &    90.44     & 86.40   &   76.75    & 87.54    \\
                               & MultiCSR*  &  {\ul 84.95}  & 89.86 & 94.42 & 89.88 & 90.33        &  84.60  &  76.52 & 87.22    \\
                               & SynCSE-r  &  84.74  & {\ul 90.15} & 94.99 &  89.82  &  \bf 90.61   &  87.80  &  \bf 77.57 &  87.95   \\
                               & MultiCSR-r  &  84.86  & \bf 90.17  & 95.00 &  89.88  & 89.68        &  88.00  & 76.29      &  87.70   \\ \midrule
\multirow{7}{*}{RoBERTa-base}  & SimCSE$\spadesuit$              & 81.04 & 87.74 & 93.28 & 86.94 & 86.60 & 84.60 & 73.68 & 84.84          \\
                               & DiffCSE$\spadesuit$             & 82.42 & 88.34 & 93.51 & 87.28 & 87.70 & 86.60 & 76.35 & 86.03          \\
                               & PromptBERT$\clubsuit$             &     83.82    & 88.72           &      93.19    &     \bf 90.36     &  88.08         &   {\ul 90.60}   &    76.75     &  87.36         \\
                               & PCL$\spadesuit$                 & 81.83 & 87.55 & 92.92 & 87.21 & 87.26 & 85.20 & 76.46 & 85.49           \\
                               & RankCSE$\spadesuit$             & 83.32 & 88.61 & \bf 94.03 & 88.88 & 89.07 &  \bf 90.80 & 76.46 & 87.31         \\
                               & SynCSE*  &  84.82  & \bf 91.31 & 93.18 &   {\ul 89.70}    & {\ul 90.28}        &  84.80  &   76.70    &  87.26   \\
                               & MultiCSR*  & \bf 84.99  & {\ul 91.23} & 93.07 &  89.42   & \bf 91.10 &  84.60  &   77.28  & {\ul 87.38} \\
                               & SynCSE-r  &  83.78  & 91.15 & 92.98 &   89.50    &  89.95       &  85.80  &   {\ul 77.33}    &  87.21   \\
                               & MultiCSR-r  &  {\ul 84.89}  & 90.70 & {\ul 93.62} &   89.50  & 90.06        &  85.40  & \bf 78.38      &   \bf 87.51  \\ \midrule
\multirow{5}{*}{RoBERTa-large} & SimCSE$\spadesuit$              & 82.74 & 87.87 & 93.66 & 88.22 & 88.58 & {\ul 92.00} & 69.68 & 86.11          \\
                               & PCL$\spadesuit$                 & 84.47 & 89.06 & 94.60 & 89.26 & 89.02 &  94.20 & 74.96 & 87.94       \\
                               & RankCSE$\spadesuit$   &  84.61 &89.27 & 94.47 &  89.99 & 89.73 & \bf 92.60 & 74.43 & 87.87          \\
                               & SynCSE*  &  {\ul 87.42}  & 92.21 & 94.19 &  \bf 90.82  &     91.60    &  85.00  &   76.87  & 88.30    \\
                               & MultiCSR*  & 87.05 & 91.87 & 94.07 &  {\ul 90.53}  &  91.60  & 88.80   &   78.26    &  88.88   \\
                               & SynCSE-r  &  87.24  & \bf 92.29 & \bf 94.65 &  90.52  &   \bf 92.37  &  91.40  &   \bf 79.01   & \bf 89.64    \\
                               & MultiCSR-r  &  \bf 87.45  & \bf 92.29 & {\ul 94.56} &  90.45   & {\ul 91.98}        &  90.80  & {\ul 78.61}      &  {\ul 89.45}   \\ \bottomrule
\end{tabular}%
}
\caption{Comparison of different sentence embedding models accuracy on transfer tasks. The value highlighted in bold is the best value, and the value underlined is the second-best value. ``$\spadesuit$'': results from \cite{liu2023rankcse}. ``$\clubsuit$'': results from \cite{wang2024large}. ``*'': we reproduce the results with the officially released corpus from \cite{zhang2023contrastive} and \cite{wang2024large}.}\label{t5}
\end{table*}

\end{document}